\documentclass[5p,times]{elsarticle}

\makeatletter
\def\ps@pprintTitle{%
 \let\@oddhead\@empty
 \let\@evenhead\@empty
 \let\@evenfoot\@oddfoot}
\makeatother

\usepackage{graphicx}

\usepackage{amssymb}

\usepackage{fontenc}
\usepackage{calc}
\usepackage{indentfirst}
\usepackage{lastpage}
\usepackage{ifthen}
\usepackage{amsmath}
\usepackage{setspace}
\usepackage{mathpazo}
\usepackage{booktabs}
\usepackage{etoolbox}
\usepackage{amsthm}
\usepackage{hyphenat}
\usepackage{footmisc}
\usepackage{geometry}
\usepackage{url}
\usepackage{mdframed}
\usepackage{tabto}
\usepackage{soul}
\usepackage{multirow}
\usepackage{microtype}
\usepackage{tikz}
\usepackage{hyperref}

\graphicspath{ {./images/} }

\usepackage{subcaption} 
\usepackage{float}
\usepackage{paralist}

\usepackage{enumitem}

\usepackage{fancyhdr}

\usepackage{cleveref}

\usepackage{pdfpages}

\usepackage{verbatim}

\usepackage{todonotes}
\setuptodonotes{fancyline, color=blue!30}
\usepackage{rotating}

\usepackage{adjustbox}
\usepackage{array}

\newcolumntype{R}[2]{%
    >{\adjustbox{angle=#1,lap=\width-(#2)}\bgroup}%
    l%
    <{\egroup}%
}
\newcommand*\rot{\multicolumn{1}{R{45}{1em}}}

\usepackage{listings}
\usepackage{color}
\usepackage{listings}
\usepackage{mwe}

\definecolor{dkgreen}{rgb}{0,0.6,0}
\definecolor{gray}{rgb}{0.5,0.5,0.5}
\definecolor{mauve}{rgb}{0.58,0,0.82}
\definecolor{dkblue}{rgb}{0,0,0.6}
\definecolor{backcolour}{rgb}{0.95,0.95,0.95}

\lstdefinestyle{interfaces}{
  float=tp,
  floatplacement=tbp,
  abovecaptionskip=-5pt
}

\usepackage{chngcntr}
\counterwithout{figure}{section}
\usepackage[labelfont=bf]{caption} 
\DeclareCaptionType{exam}[Example][List of Examples]

\usepackage{url}

\begin{document}

\begin{frontmatter}


\title{Why model why? Assessing the strengths and limitations of LIME}

\author{Jürgen Dieber} 

\author{Sabrina Kirrane}

\address{Institute for Information Systems and New Media, \\ Vienna University of Economics and Business, Welthandelsplatz 1, 1020 Vienna, Austria}

\begin{abstract}
When it comes to complex machine learning models, commonly referred to as black boxes, understanding the underlying decision making process is crucial for domains such as healthcare and financial services, and also when it is used in connection with safety critical systems such as autonomous vehicles. As such interest in explainable artificial intelligence (xAI) tools and techniques has increased in recent years. However, the effectiveness of existing xAI frameworks, especially concerning algorithms that work with data as opposed to images, is still an open research question. In order to address this gap, in this paper we examine the effectiveness of the Local Interpretable Model-Agnostic Explanations (LIME) xAI framework, one of the most popular model agnostic frameworks found in the literature, with a specific focus on its performance in terms of making tabular models more interpretable. In particular, we apply several state of the art machine learning algorithms on a tabular dataset, and demonstrate how LIME can be used to supplement conventional performance assessment methods. In addition, we evaluate the understandability of the output produced by LIME both via a usability study, involving participants who are not familiar with LIME, and its overall usability via an assessment framework, which is derived from the International Organisation for Standardisation 9241-11:1998 standard.
\end{abstract}

\begin{keyword}
Machine learning \sep Explainable Artificial Intelligence \sep Model Agnostic Explanations

\end{keyword}

\end{frontmatter}

\section{Introduction}\label{sec:Sec1} 

Since the term was first mentioned in 1956 \cite{History}, artificial intelligence (AI), and especially its subset machine learning, has steadily made its way into various kinds of industries and aspects of our lives, like healthcare\footnote{https://www.entrepreneur.com/article/341626}\footnote{https://medicus.ai/de/}, transportation\footnote{https://kodiak.ai/} and advertisement\footnote{https://instapage.com/blog/machine-learning-in-advertising}\footnote{https://www.ezoic.com/}. While machine learning applications are advancing further, the understanding of how machine learning models work and how decisions are made is not advancing at the same pace. In some applications like recommendation systems or predictive maintenance it may not be necessary to understand the black box decision making, as long as the models' predictions are accurate in the majority of cases. However, in circumstances where human lives are involved, like medical diagnosis or self-driving cars, the ability to understand the decision process is essential in order to establish trust in such systems. 

Explainable AI (xAI) \cite{koh2017understanding} is a generic term used to describe the intention to build and use models that can be interpreted and understood by their users. One approach is to develop powerful and fully explainable models, such as deep k-nearest neighbours \cite{papernot2018deep} and teaching explanations for decisions \cite{hind2018ted}. Another approach is to tackle the issue of model agnostic post modelling interpretability, by explaining the output of well established machine learning models, instead of replacing these models entirely, cf., SHAP by \citet{lundberg2017unified}, LIME by \citet{Ribeiro} and MAPLE by \citet{plumb2018model}. 

When it comes to xAI frameworks, the Local Interpretable Model-Agnostic Explanations (LIME) framework is one of the predominant tools discussed in the literature, especially when it comes to its image explainer \cite{lundberg2017unified,guidotti2018local,peltola2018local, Sokol_2020}, however in contrast its tabular explainer has rarely been evaluated to date \cite{Sokol_2020}. Additionally, existing works have primarily used LIME as a benchmarking framework as opposed to assessing the effectiveness of its output \cite{hu2018locally, Mishra2017LocalIM, zafar2019dlime}. Thus the overarching goal of this paper is to evaluate the effectiveness of the output produced by LIME, when it comes to machine learning algorithms that work with data as opposed to images.
Summarizing our contributions, we:
\begin{inparaenum}[(i)]
  \item demonstrate how LIME can be used to supplement conventional performance assessment methods;
  \item evaluate the understandability of the output produced by LIME via a usability study; and
  \item propose an assessment framework, which is derived from the International Organisation for Standardisation (ISO) 9241-11:1998 standard, that can be used not only to evaluate the usability of LIME but also other xAI frameworks.
\end{inparaenum}
In addition, our code and data are made available in a GitHub repository\footnote{https://github.com/jdieber/WhyModelWhy}.

The reminder of this article is structured as follows: \Cref{sec:Sec2} summarizes the state of the art with respect to post-modelling interpretability. In \Cref{sec:Sec3} we compare the performance of several machine learning models using conventional methods. In \Cref{sec:Sec4} we illustrate the value LIME adds when it comes to understating the models output in comparison to conventional performance assessment methods. \Cref{sec:Sec5} examines the interpretability of the output produce by LIME. Finally, we present our conclusions and interesting directions for future work in \Cref{Sec:Sec6}. 

\begin{table*}[t!]
\centering
\small
\scalebox{0.9}{
\begin{tabular}{|l|l|l|l|l|}
\hline


\vtop{\hbox{\strut} \hbox{\strut \textbf{Method}}\hbox{\strut}}  &

\vtop{\hbox{\strut} \hbox{\strut 
\textbf{Reference}}\hbox{\strut}}  & 

\vtop{\hbox{\strut} \hbox{\strut 
\textbf{Scope}}\hbox{\strut}}  & 

\vtop{\hbox{\strut} \hbox{\strut 
\textbf{Data Type}}\hbox{\strut}}  & 

\vtop{\hbox{\strut} \hbox{\strut 
\textbf{Evaluation Technique }}\hbox{\strut}} \\

\hline
    
\textit{Activation maximization} & \cite{nguyen2016synthesizing} & Global & Image & -    \\\hline

\textit{Counterfactual} & \cite{karimi2019model}, \cite{sharma2019certifai} & Local & Tabular, Image & -   \\\hline

\textit{Feature importance} & \cite{lei2018distribution}, \cite{casalicchio2018visualizing} & \vtop{\hbox{\strut Global,} {\hbox{\strut Local}}} & Image, Tabular & -  \\\hline

\textit{Fisher kernels} & \cite{khanna2018interpreting} & Local & Image &
\vtop{
{\hbox{\strut \textit{Baseline evaluation: }}
{\hbox{\strut Fisher kernels compared}
{\hbox{\strut to Influence functions}}}
}}

\\\hline

\textit{Frequency map} & \cite{dhurandhar2019model} & Local & Tabular &    \vtop{
{\hbox{\strut \textit{Baseline evaluation: }}
{\hbox{\strut MACEM compared to LIME}}
\hbox{\strut \textit{User interview:} }}
{\hbox{\strut MACEM compared to LIME}}}\\\hline

\textit{if-then rules} & \cite{ribeiro2016nothing}, \cite{puri2017magix} & \vtop{\hbox{\strut Global,} {\hbox{\strut Local}}} & Image, Tabular, Text & -   \\\hline

\textit{Influence function} & \cite{koh2017understanding} & Local & Image, Tabular, Text & - \\\hline

\textit{LIME} & \cite{Ribeiro}, \cite{katuwal2016machine}, \cite{ribeiro2016nothing}, \cite{Sokol_2020}  & \vtop{\hbox{\strut Global,} {\hbox{\strut Local}}} & Image, Tabular, Text & -   \\\hline

\textit{LIME extension} &   \vtop{\hbox{\strut \cite{singh2019exs}, \cite{guidotti2018local},  \cite{hu2018locally},}\hbox{\strut \cite{zafar2019dlime}, \cite{peltola2018local}, \cite{Mishra2017LocalIM}} }& Local & Image, Tabular, Text & 
\vtop{\hbox{\strut \textit{Baseline evaluation:} SUP-LIME}
{\hbox{\strut compared to K-LIME;}}
{\hbox{\strut SLIME compared to}}
{\hbox{\strut positive saliency map;} }
{\hbox{\strut DLIME compared to LIME}}}
\\\hline

\textit{MAPLE} & \cite{plumb2018model} & \vtop{\hbox{\strut Global,} {\hbox{\strut Local}}} & Tabular &   
\vtop{\hbox{\strut \textit{Baseline evaluation:} \textit{MAPLE}} {\hbox{\strut compared to LIME}}}\\\hline

\textit{Model distillation} & \cite{tan2017detecting} & Global & Tabular & -   \\\hline

\textit{Parametric statistical tests} & \cite{garcia2009study} & Local & Tabular & -   \\\hline

\textit{Partial dependence plot} & \cite{green2010modeling} & \vtop{\hbox{\strut Global,} {\hbox{\strut Local}}} & Tabular & -   \\\hline

\textit{Prototype and criticism} & \cite{elith2008working} & \vtop{\hbox{\strut Global,} {\hbox{\strut Local}}} & Tabular & -   \\\hline

\textit{Ranking models} & \cite{singh2020model} & Local & Text & - \\\hline

\textit{Relevance scores} & \cite{arras2017relevant} & Local & Text &    
\vtop{\hbox{\strut \textit{Baseline evaluation:} LRP} {\hbox{\strut compared to TFIDF and uniform}}}\\\hline

\textit{Saliency map} & \vtop{\hbox{\strut \cite{baehrens2010explain}, \cite{zeiler2014visualizing}, \cite{zhou2016learning}, }\hbox{\strut \cite{sundararajan2017axiomatic}, \cite{fong2017interpretable}, \cite{dabkowski2017real} }} & Local & Tabular, Text, Image & - \\\hline

\textit{Sensitive analysis} & \cite{cortez2011opening} & \vtop{\hbox{\strut Global,} {\hbox{\strut Local}}} & Tabular & -  \\\hline

\textit{Shapley value} &  \vtop{\hbox{\strut \cite{lundberg2017unified}, \cite{casalicchio2018visualizing}, \cite{lundberg2016unexpected}, }\hbox{\strut \cite{chen2018shapley}, \cite{frye2019asymmetric}  }}
& Local & Tabular, Text, Image & 
\vtop{\hbox{\strut \textit{Baseline evaluation:} true}
{\hbox{\strut shapley value, classical shapley}}
{\hbox{\strut estimations, LIME and ES values}}

{\hbox{\strut \textit{User interview:} SHAP}}
{\hbox{\strut compared to true shapley Value,}}
{\hbox{\strut LIME and shapley sampling}}} 
\\\hline

\textit{Surrogate models} & \cite{Ribeiro}, \cite{bastani2017interpretability}, \cite{thiagarajan2016treeview} & \vtop{\hbox{\strut Global,} {\hbox{\strut Local}}} & Image, Tabular, Text  &  -  \\\hline

\textit{Visualisation} & \cite{casalicchio2018visualizing} & \vtop{\hbox{\strut Global,} {\hbox{\strut Local}}} & Tabular & -   \\\hline

\end{tabular}}
\caption{Existing model agnostic explainablility approaches}
\label{Tab:Review} 
\end{table*}

\section{A comparative analysis of existing work on model agnostic explainablility}\label{sec:Sec2} 

Generally speaking existing work relating to xAI can be grouped into two distinct categories: (i) the development of fully explainable models, cf., \cite{papernot2018deep, hind2018ted} and (ii) the development of model agnostic explainablility frameworks, cf., \cite{lundberg2017unified, Ribeiro, plumb2018model}. Considering that model agnostic frameworks can be used with any machine learning algorithm in this paper we focus specifically on the latter. In particular, we compare and contrast existing work with respect to the scope of the interpretability, the type of data the method is tested with, and the evaluation used to assess or compare the methods performance.

In terms of the scope of interpretability, a framework can either be on a \textit{global level}, meaning it makes different models comparable with each other, by summarizing their performance with respect to specific indicators, or on a \textit{local level}, giving insight into how a classification in the case of a single prediction is made. Although the vast majority of works focus on local interpretability \cite{arras2017relevant,baehrens2010explain,bastani2017interpretability,casalicchio2018visualizing,cortez2011opening,dabkowski2017real,dhurandhar2019model,elith2008working,fong2017interpretable,frye2019asymmetric,garcia2009study,green2010modeling,guidotti2018local,hu2018locally,karimi2019model,katuwal2016machine,khanna2018interpreting,koh2017understanding,lei2018distribution,lundberg2016unexpected,lundberg2017unified,Mishra2017LocalIM,peltola2018local,plumb2018model,Ribeiro,ribeiro2016nothing,sharma2019certifai,singh2019exs,singh2020model,sundararajan2017axiomatic,thiagarajan2016treeview,zafar2019dlime,zeiler2014visualizing,zhou2016learning}, several can also be used for a global comparison \cite{bastani2017interpretability,casalicchio2018visualizing,cortez2011opening,elith2008working,green2010modeling,lakkaraju2017interpretable,nguyen2016synthesizing,lei2018distribution,plumb2018model,puri2017magix,Ribeiro,tan2017detecting,thiagarajan2016treeview}. Only the activation maximization method \cite{nguyen2016synthesizing} and model distillation \cite{tan2017detecting} are exclusively global.
Although each of the papers includes some demonstration of the method using a specific data type, the actual data used is very different: twenty-four methods are applied to tabular data \cite{baehrens2010explain,bastani2017interpretability,casalicchio2018visualizing,cortez2011opening,dhurandhar2019model,elith2008working,frye2019asymmetric,garcia2009study,green2010modeling,hu2018locally,karimi2019model,katuwal2016machine,koh2017understanding,lei2018distribution,lundberg2016unexpected,Mishra2017LocalIM,plumb2018model,puri2017magix,Ribeiro,ribeiro2016nothing,sharma2019certifai,sundararajan2017axiomatic,tan2017detecting,zafar2019dlime}, sixteen are applied to image data \cite{baehrens2010explain,dabkowski2017real,fong2017interpretable,guidotti2018local,khanna2018interpreting,koh2017understanding,lundberg2017unified,nguyen2016synthesizing,peltola2018local,sharma2019certifai,sundararajan2017axiomatic,thiagarajan2016treeview,zeiler2014visualizing,zhou2016learning}, and eight are applied to textual data \cite{arras2017relevant,koh2017understanding,lundberg2016unexpected,Ribeiro,ribeiro2016nothing,singh2019exs,singh2020model,sundararajan2017axiomatic}. Only four publications, \citet{koh2017understanding}, \citet{Ribeiro, ribeiro2016nothing} and \citet{sundararajan2017axiomatic} include an application of all three data types.

Concerning the evaluation technique, where an assessment is performed two different methods are used: a \textit{baseline evaluation} and a \textit{user interview}. A baseline evaluation is a quantitative evaluation technique, where one or more indicators are used for a comparative assessment. For instance, Plumb et al. \cite{plumb2018model} uses a self defined causal local explanation metric to compare their framework to LIME. In total, eight of the publications apply some sort of baseline evaluation \cite{arras2017relevant,dhurandhar2019model,hu2018locally,khanna2018interpreting,lundberg2016unexpected,Mishra2017LocalIM,plumb2018model,zafar2019dlime}.
The second evaluation technique is a qualitative method, either a survey or user interview. Only three publications use this approach. \citet{lakkaraju2017interpretable} and \citet{lundberg2017unified} include a survey in their evaluation and \citet{dhurandhar2019model} ask two professionals to rate a mixed set of interpretability framework outputs given to them. Out of the ten publications who evaluate their framework, six draw a comparison to LIME \cite{dhurandhar2019model,hu2018locally,lundberg2016unexpected,lundberg2017unified,plumb2018model,zafar2019dlime}, from which we can assume that LIME constitutes a benchmark for interpretability frameworks. However, when it comes to the evaluation of LIME itself, none of the publications actually use evaluation techniques to assess LIMEs performance and only \citet{Sokol_2020} evaluate LIME as a demonstration of their novel explainability taxonomy.

Model agnostic frameworks have also been applied in several domains. Within the medical sector, considering that AI systems are used to support the diagnosis, both \citet{gale2018producing} and \citet{katuwal2016machine} identify the need to enhance model comprehensibility for the professionals using them. In the case of Holzinger et al. 2019 \cite{doi:10.1002/widm.1312} they go beyond simply explaining the models, towards uncovering causality. Within the field of news detection, the automatic understanding or processing of text, xAI helps to shed light on the multi-layer deep learning applications used for advanced applications \cite{arras2017relevant}. While, in the music business, content analysis is supported by model agnostic interpretability frameworks in order to gain a better understanding of how certain tones are identified \cite{Mishra2017LocalIM}. 

Although the LIME framework, especially its image explainer, is one of the predominant tools discussed in the literature, its tabular explainer has received limited attention to date. In addition, existing work focuses primarily on using LIME as a benchmark as opposed to assessing the usability of LIME itself. In order to fill this gap in this paper we apply LIME on tabular machine learning models and evaluate LIMEs performance in terms of comparability, interpretability and usability.


\section{Using machine learning to classify tabular data}\label{sec:Sec3} 


We start by presenting four state of the art classification models, namely decision trees, random forest, logistic regression and XGBoost. Following on from this, we make use of conventional methods (i.e., the classification report and receiver operating characteristic curve) to assess the model performance and identify the best performing algorithm.

	\subsection{Tabular data pre-processing}

For our tabular data analysis we use the \textit{Rain in Australia} data-set from Kaggle\footnote{https://www.kaggle.com/jsphyg/weather-dataset-rattle-package}. Before the algorithm is trained, we work through the different variables step by step to fully understand their meaning and make them processible by our model. Then the dependencies amongst each other are evaluated and those that could potentially bias the models or have too many missing values are removed. A summary of the full dataset is given in \Cref{Tab:data}, while the features we use for training are denoted with a star. 

\begin{table*}[t!]
\centering
\small
\scalebox{0.80}{
\begin{tabular}{|p{3.2cm}|p{3.2cm}|p{3.2cm}|p{3.2cm}|p{3.2cm}|}
 \hline
\textbf{variable name} & \textbf{sample input} & \textbf{type} & \textbf{non-null-values}\\
 \hline
Date & 2008-12-03 & categorical & 142193  \\ 
Location & Albury & categorical & 142193  \\ 
MinTemp*  & 13.4 & numerical & 141556  \\ 
MaxTemp* & 25.1 & numerical & 141871  \\ 
Rainfall* & 0.00 & numerical & 140787  \\ 
Evaporation & 23 & numerical & 81350  \\ 
Sunshine & 11 & numerical & 74377  \\ 
WindGustDir* & W & categorical & 132863  \\ 
WindGustSpeed* & 44.0 & numerical & 132923  \\ 
WindDir9am* & NW & categorical & 132180  \\ 
WindDir3pm* & W & categorical & 138415  \\ 
WindSpeed9am* & 25.0 & numerical & 140845  \\ 
WindSpeed3pm* & 8.0 & numerical & 139563  \\
Humidity9am* & 25.0 & numerical & 140419  \\ 
Humidity3pm* & 22.0 & numerical & 138583  \\ 
Pressure9am* & 1007.7 & numerical & 128179  \\ 
Pressure3pm* & 1007.1 & numerical & 128212  \\ 
Cloud9am & 2.0 & numerical & 88536  \\ 
Cloud3pm & 8.0 & numerical & 85099  \\ 
Temp9am* & 16.9 & numerical & 141289  \\ 
Temp3pm* & 21.8 & numerical & 139467  \\ 
RainToday* & Yes & categorical & 140787  \\ 
RISK\_MM & 0.2 & numerical & 142193  \\ 
RainTomorrow* & No & categorical & 142193  \\ 
 
\hline
\end{tabular}}
\caption{An overview of the datasets' features}
\label{Tab:data} 
\end{table*}

From a preprocessing perspective, we modify several categorical variables, making them numeric so they can be processed by the models. We further build a scikit learn \texttt{pipeline} object, to apply the preprocessor on the data and sequentially build our model based on its structure. This enables us to perform a sequence of different transformations and to give each algorithm a customised setting while being able to cross-validate each setting-combination during the training process. 

The scikit-learn \texttt{train\_test\_split} function is used to break our data into different parts, namely training and testing data. We assign 70\% of our observation to the training dataset and the remaining 30\% to the testing dataset. Once the data is prepared, we train our four models with the same training data. For comparability reasons, we mainly used standard parameter settings for the setup of the algorithms.

	    \subsection{The application and interpretation of the machine learning models}
	    
Inorder to analyse the models performance on the testing data, we utilise the the sklearn classification report. A model comparison using these conventional methods is presented in  \Cref{Tab:High_Class}.
Precision, recall and f1-score are calculated based on the classification results true positive, true negative, false positive and false negative. True positive and true negative both indicate that the weather was correctly predicted with either it is going to rain or it is not going to rain, respectively. A false positive however indicates a class that should not have been predicted positive and false negative indicates that a class should have been predicted positive. The scores next to the metrics name in \Cref{Tab:High_Class} either refer to the target variable that it is not going to rain (0) or that it is going to rain (1) as well as the weighted scores (w) and the training baseline value (tr) for the receiver operating characteristic curves (ROC). Taking the decision tree as an example, the values are then calculated as follows:

\begin{description}[style=unboxed]

    \item[Accuracy:] The accuracy gives an average of how often the model classified the target variable correctly, in the decision trees example in 79\% of the time. 
    \item[Precision:] The precision describes how often the model was correct in classifying an observation as positive. It is the result of the true positives, divided by the sum of false positives and true positives, adding up to 91\% for the outcome that it is not going to rain and 53\% for the outcome that it is going to rain.
    \item[Recall:] For the recall measurement, the performance of the variables is more similar. It consists of the true positives divided by the sum of true positives and false negatives, 81\% and 73\%, respectively. 
    \item[F1-score:] The f1-score tells us what percentage of positive prediction is correct, including the recall and precision into its measurement. The f1-score consists of two times the precision * recall divided by the sum of precision and recall. The decision tree delivers a f1-score of 86\% for the outcome that it is not going to rain and 61\% for it is going to rain.
    \item[Macro score:] The macro score represents the overall performance of the indicator, meaning the average. The macro precision reaches 82\%, the macro recall 71\% and the macro f1-score 74\%. 
    \item[Weighted average score:] The weighted average is the respective score times its number of instances, for example, the 0.85\% weighted average precision result from the target variable not going to rain, having a score of 91\% and 53\% of target variable going to rain, respectively.
\end{description}

\begin{figure*}[t!]
\centering
\begin{subfigure}{0.245\textwidth}
\includegraphics[width=1.0\linewidth]{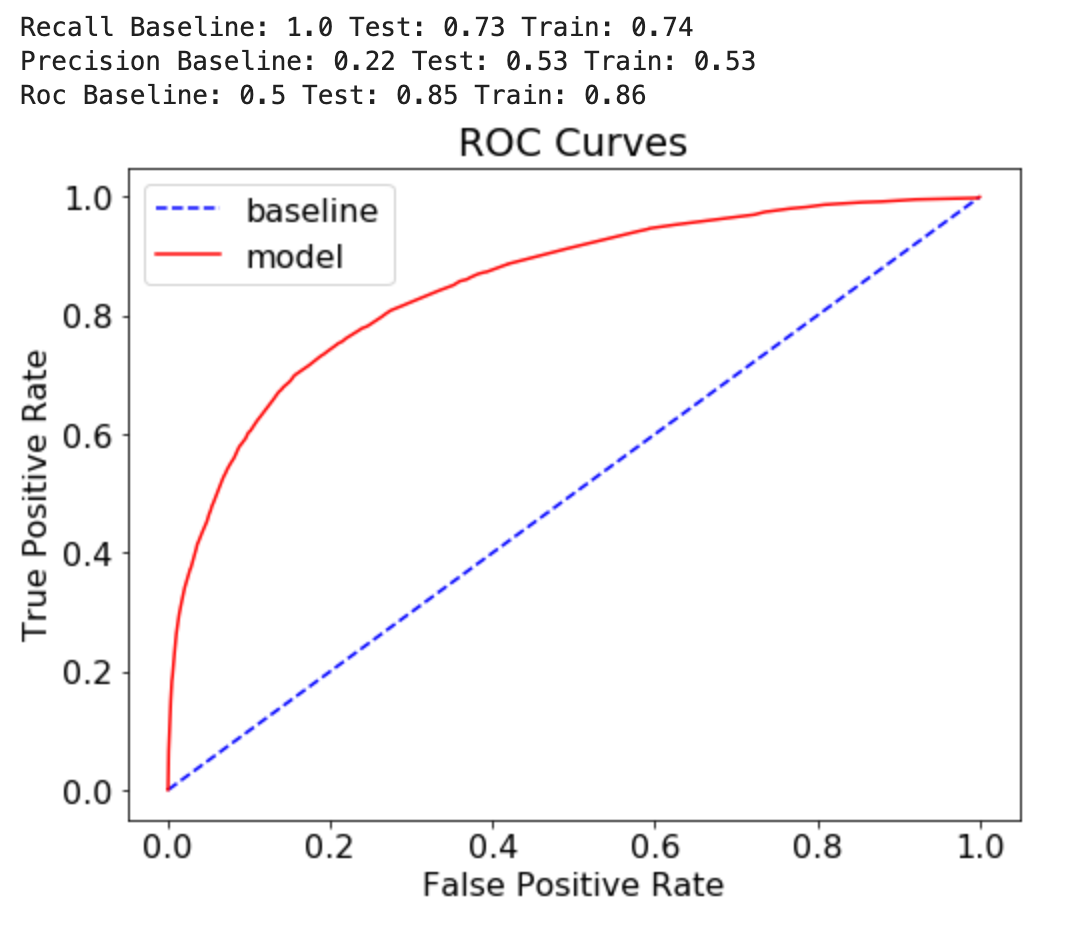} 
\caption{Decision Tree}
\label{fig:subim1}
\end{subfigure}%
\begin{subfigure}{0.245\textwidth}
\includegraphics[width=1.0\linewidth]{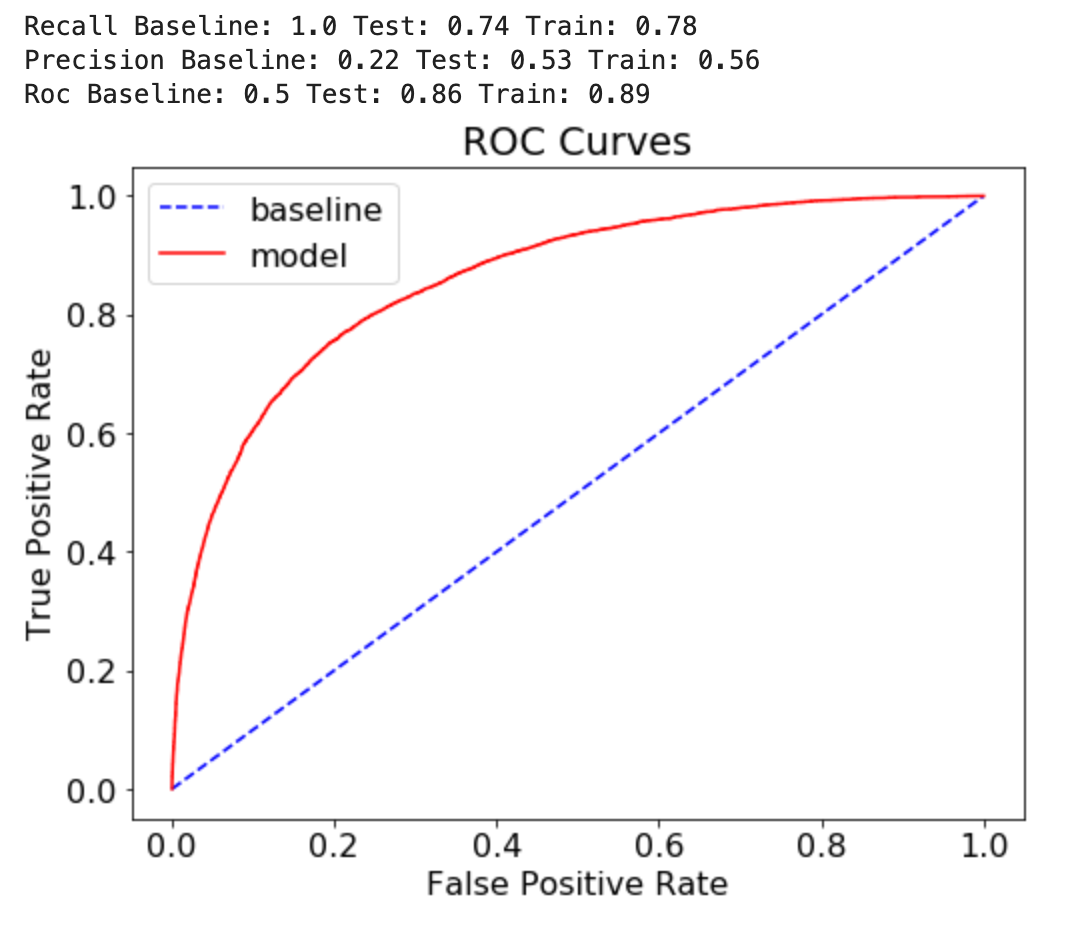}
\caption{Random Forest}
\label{fig:subim2}
\end{subfigure}%
\begin{subfigure}{0.245\textwidth}
\includegraphics[width=1.0\linewidth]{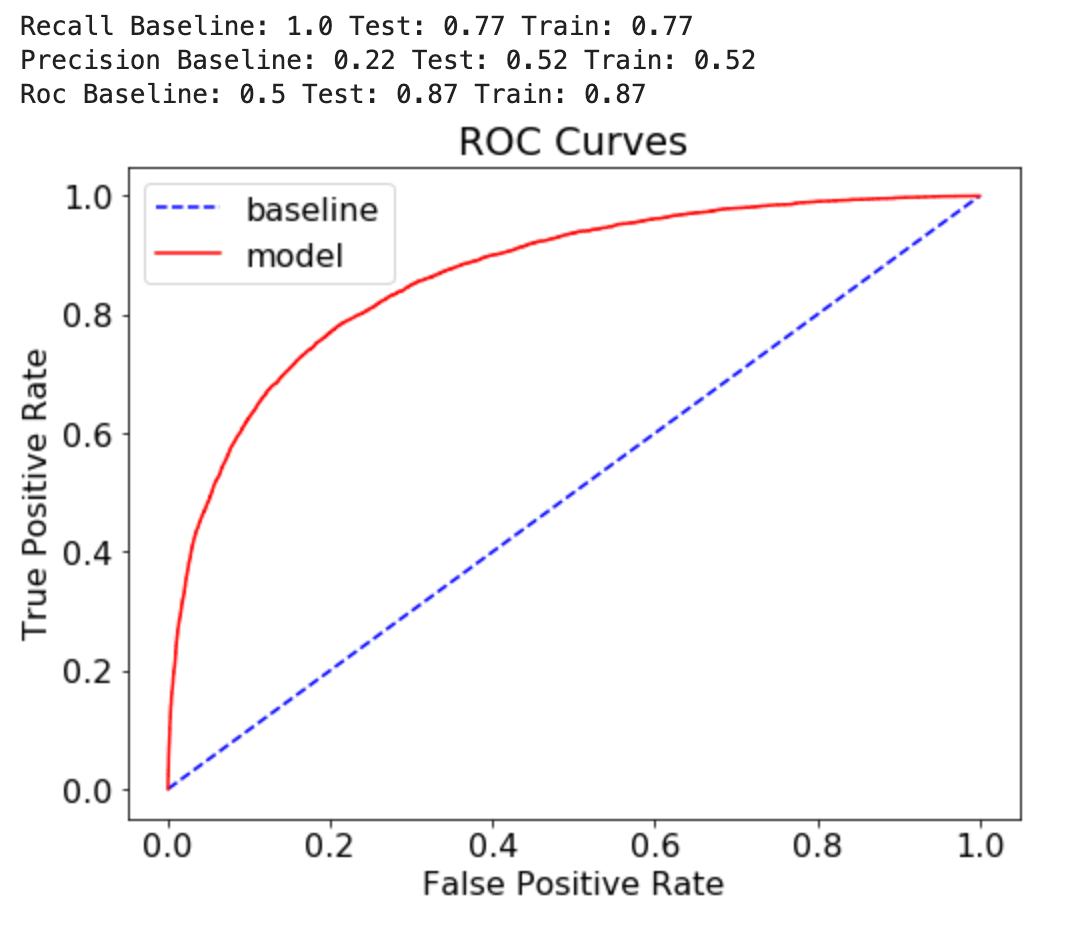}
\caption{Logistic Regression}
\label{fig:subim3}
\end{subfigure}%
\begin{subfigure}{0.245\textwidth}
\includegraphics[width=1.0\linewidth]{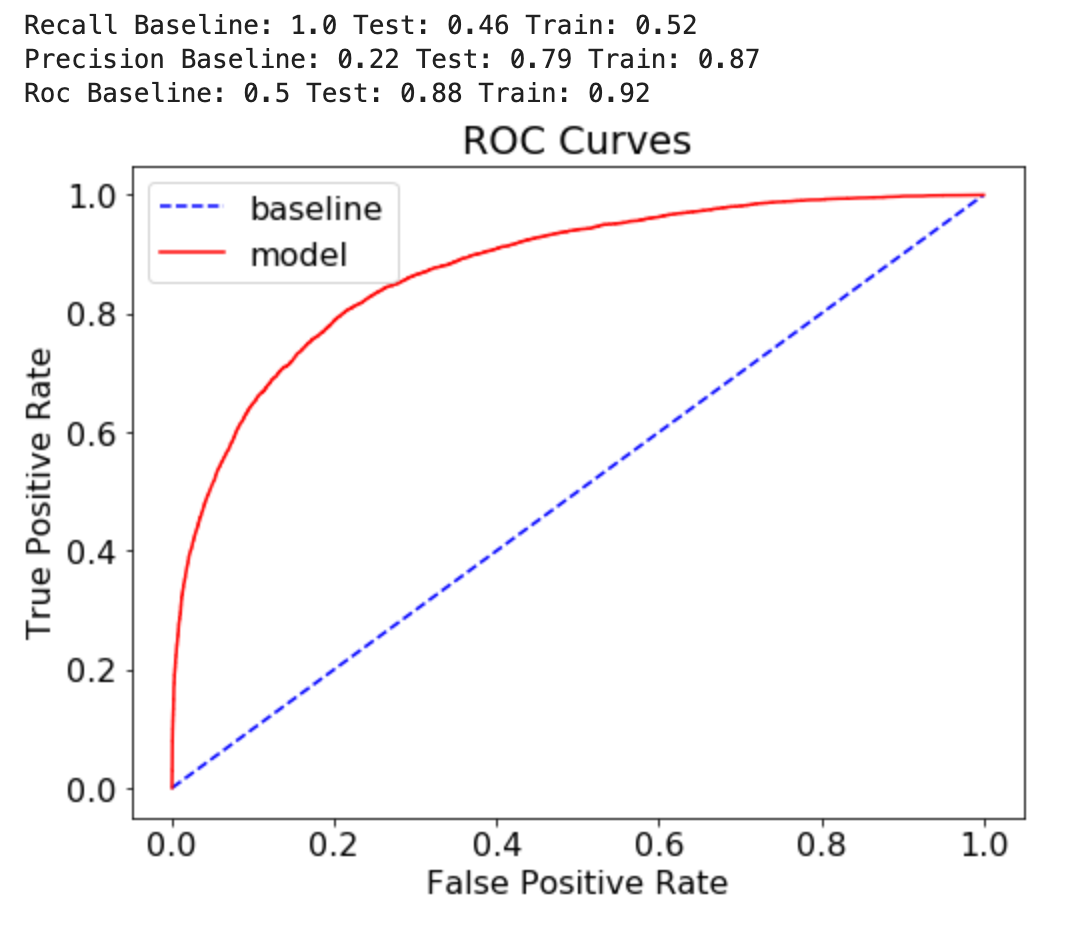}
\caption{XGB}
\label{fig:subim4}
\end{subfigure}%

\caption{The ROC curves of the models}
\label{fig:ROC_Collection}
\end{figure*}

Another state of the art tool to measure the validity of classification results is the ROC \cite{ROC}. \Cref{fig:ROC_Collection} displays one ROC curve per model, each graph showing two curves, the upper one is the ROC curve, posing a probability, the lower one is the \textit{baseline}, which separates the ROC and the area under the curve (AUC), which is a measurement for separability. Similar to the \texttt{classification report}, the ROC curve uses \textit{precision} and \textit{recall} for its measurement but due to its graphical display curves of different models can be easily compared with each other.
The further to the upper left corner the curve bends, the better the classification. The AUC measures the general accuracy, meaning how well a model can differentiate between classes. For the AUC the following rule holds true: the closer its value is to 1, the better the model is able to correctly classify. If the value is 0.5 it means that the model is not better than randomly guessing and a value of close to 0 means that the model is doing the classification upside down\footnote{https://machinelearningmastery.com/roc-curves-and-precision-recall-curves-for-classification-in-python/ }\footnote{https://www.jstor.org/stable/2531595?seq=1}. 

\begin{table*}[t!]
\centering
\small
\scalebox{0.9}{
\begin{tabular}{rccccccccccc}

\rot{model} &
\rot{accuracy} & 
\rot{precision (0)} &  
\rot{precision (1)} & 
\rot{recall (0)} & 
\rot{recall (1)} & 
\rot{f1-score (0)} & 
\rot{f1-score (1)}  & 
\rot{precision (w)} &
\rot{recall (w)} &
\rot{f1-score (w)} &
\rot{ROC (tr)}
\\\toprule
    
decision tree & 0.79 & 0.91 & 0.53 & 0.81 & 0.73 & 0.86 & 0.61 & 0.83 & 0.79 & 0.80 & 0.85 \\\hline
random forest & 0.80 & \textbf{0.92} & 0.53 & 0.81 & 0.75 & 0.86 & \textbf{0.62} & 0.83 & 0.80 & 0.81 & 0.86 \\\hline
logistic reg. & 0.79 & \textbf{0.92} & 0.52 & 0.80 & \textbf{0.77} & 0.86 & \textbf{0.62} & 0.84 & 0.79 & 0.80 & 0.87 \\\hline
XGBoost & \textbf{0.85} & 0.86 & \textbf{0.79} & \textbf{0.96} & 0.46 & \textbf{0.91} & 0.58 & \textbf{0.85} & \textbf{0.85} & \textbf{0.84} & \textbf{0.88} \\ \bottomrule
\end{tabular}}
\caption{A model comparison using conventional methods}
\label{Tab:High_Class} 
\end{table*}

    \subsection{An assessment of the machine learning models} 

%
Overall it is notable that the performances of the decision tree, random forest and logistic regression are very similar while the XGBoost performance differs significantly. 
In this comparison, the XGBoost delivers the highest values with a 85\% \textit{accuracy}, \textit{weighted average scores} of 85\% \textit{precision}, 85\% \textit{recall} as well as 84\% \textit{f1-score}. But it's weak performance in classifying that it is going to rain correctly, can be seen in a low \textit{recall (1)} and \textit{f1-score (1)} score with 46\% and 58\%, respectively. It is worth noting that the high difference in the \textit{recall} scores for the respective target variable might be caused by unbalanced testing data, which is something we would like to further explore in future work. 
The logistic regression offers the highest \textit{recall (1)}, in the case of 77\% of the positive observations it predicts correctly that it is going to rain, with a weighted \textit{recall} of 79\%. In terms of \textit{f1-score (1)} the logistic regression and the random forest score equal 62\% which is four percent higher than the XGBoost with 58\%.
Furthermore, comparing the ROC curves shows a similar performance for all models, with XGBoost scoring 88\% \textit{ROC baseline}, the logistic regression 87\%, the random forest 86\% and the decision tree 85\%.

To summarize, the decision tree performs worst in all metrics. The random forest and the logistic regression never differ more than two percent in any of the metrics and are therefore performing similarly. Although XGBoost outperforms the others in several metrics, it scores significantly lower when it comes to predicting the outcome of a positive observation. Thus, in order to decide which model should be deployed, based on this results, requires a trade-off: a higher accuracy and more accurate prediction of true negatives would stand in favor of the XGBoost, while the need for a more accurate prediction of true positives would stand in favor of the random forest or the logistic regression. 
%
Furthermore, while the confusion matrix and the ROC give us insight into how the models perform, they do not reveal how the models reach a certain decision.

\section{Applying the LIME xAI framework to tabular data}\label{sec:Sec4}


In order to better understand the behaviour of our four classification models we employ the Local Interpretable Model-Agnostic Explanations (LIME) xAI framework. We start by providing a short introduction to LIME. Following on from this, we apply LIME on our four tabular models and describe the output. Finally, we conduct a quantitative analysis of fifty aggregated LIME observations to further compare performance on a global level.

    \subsection{A short introduction to LIME}

LIME is an open source framework, published by Ribeiro et al. in 2016 \cite{Ribeiro}, which aims to shed light on the decision-making process of machine learning models and therewith establish trust in their usage. \textit{Local} means that the framework analyses specific observations. It does not give a general explanation as to why the model behaves in a certain way, but rather explains how a specific observation is categorised. 
\textit{Interpretable} means that the user should be able to understand what a model does. Thus, in image classification it shows which part of the picture it considered when it comes to predictions and when working with tabular data it shows which features influence its decision. \textit{Model-Agnostic} means that it can be applied to any blackbox algorithm we know today or that we might develop in the future. If the model is a glassbox this is not taken into consideration as LIME treats every model like a blackbox. \textit{Explanations} denote the output, which the LIME framework produces. LIME has three core functionalities: the image explainer interprets image classification models, the text explainer provides insight into text based models\footnote{https://www.tensorflow.org/lite/models/text\_classification/overview} and the tabular explainer assesses to what extent features of a tabular dataset are considered when it comes to the classification process\footnote{https://towardsdatascience.com/pytorch-tabular-binary-classification-a0368da5bb89}. 

	\subsection{The application of the LIME Tabular Explainer}

\begin{lstlisting}[style=interfaces, language=Python, label={lst:explainer}, caption = The LIME tabular explainer]
explainer = LimeTabularExplainer(
    convert_to_lime_format(X_train, categorical_names).
    values,
    mode="classification",
    feature_names=X_train.columns.tolist(),
    categorical_names=categorical_names,
    categorical_features=categorical_names.keys(),
    discretize_continuous=True,
    random_state=42)
                        
\end{lstlisting}

The main function LIME offers, called \texttt{explainer}, enables us to call a specific observation and get an interpretation as a result. The following steps demonstrate its deployment.

\begin{description}[style=unboxed]

\item[The convert to LIME function:]
Prior to being able to explain an observation, we need to convert the output into a certain format, which we do by creating a list of all possible categorical values per feature. Then, we use the \texttt{convert\_to\_lime\_format} function \cite{KleMag} adopted from Kevin Lemagnen's Pycon presentation in 2019\footnote{https://speakerdeck.com/klemag/pycon-2019-introduction-to-model-interpretability-in-python}, as the one included in the LIME documentation only works with older versions of Python. The function converts all existing string variables to integers, such that they can be interpreted.

\item[The explainer:]
The explainer itself is included in the LIME library and displayed in \autoref{lst:explainer}. We set all parameters manually, as the explainer does not possess any default values. First we call our now formatted dataset and set the mode to classification, then we give a list of all features in our dataset (line 3) and with \texttt{categorical\_names=categorical\_names} we specify which of the variables are categorical (line 4), \texttt{Categorical\_features} (line 5) lists the index of all features with a categorical type and \texttt{discretize\_continuous} (line 6) is a mathematical function that simply helps to produce a better output by converting continuous attributes to nominal attributes. The final parameter, \texttt{random\_state}, brings consistency into the function, otherwise it always picks a different number whenever we reload the function.

\item[Displaying one observation:]
We choose one observation on which we apply the interpretability framework and subsequently print the classification that each model gives for this instance as well as the true label. We can now convert the output to the LIME format, saving it in the observation variable before defining a standard predict function. The \texttt{custom\_predict\_proba} function, is able to transform very simple models but also more complex input. It converts the data so that it is processible by the \texttt{LIMETabularExplainer}, which we carry out for every model we wish to interpret. After this we can apply the LIME framework on our classification models. To create a LIME output, we define the explanation as  \texttt{explainer.explain\_instance} and include the observation we chose above, adding the \texttt{lr\_predict\_proba} and five features as this shows us the factors considered the most influential on predicting the target variable.

\end{description}

\begin{figure*}[t!]
\includegraphics[width=0.9\textwidth]{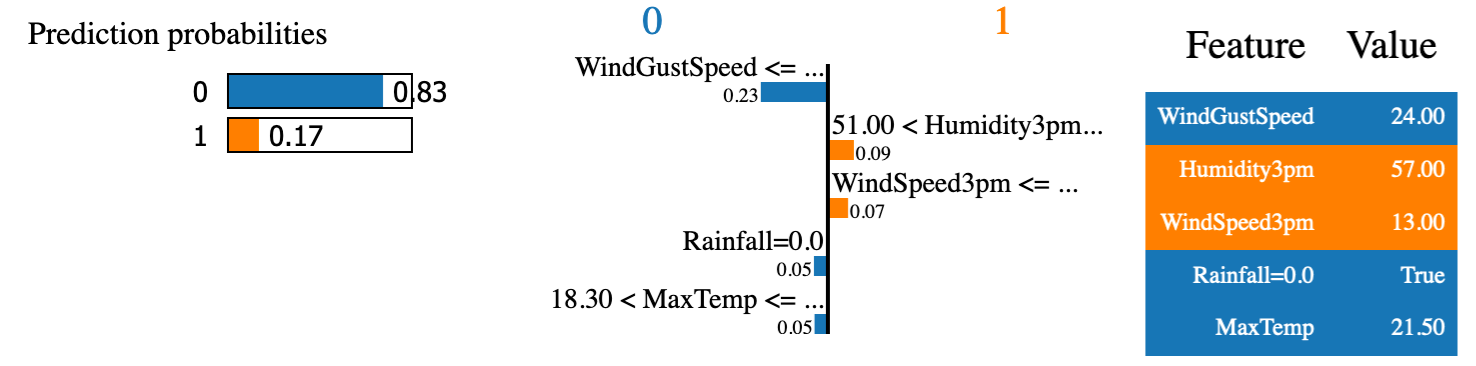}
\caption{Example LIME output using logistic regression}
\label{fig:LIME}
\end{figure*}

\begin{figure*}[t!]
\centering
\includegraphics[width=0.8\linewidth]{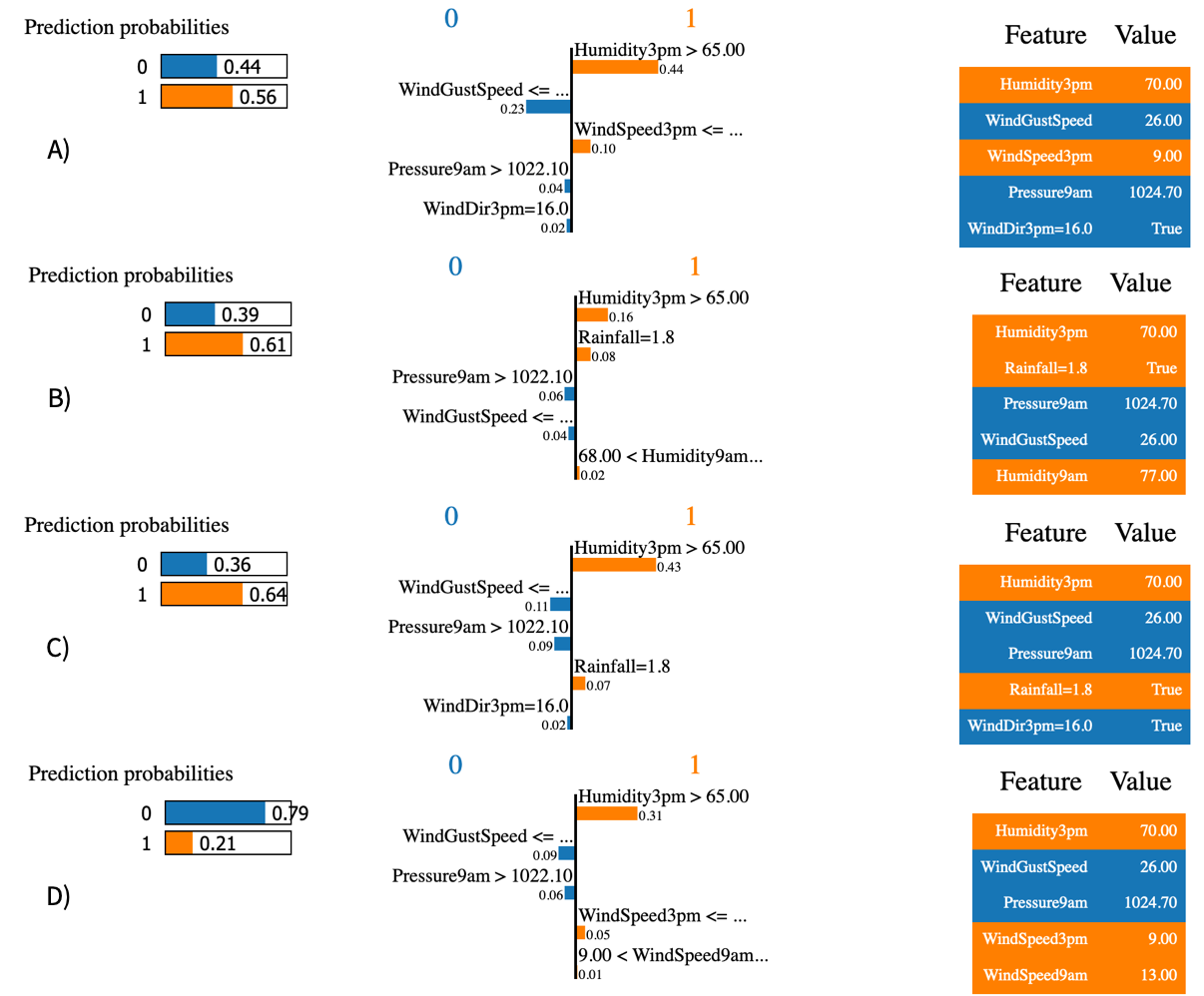}
\caption{LIME output of the same observation from the (A)Decision Tree, (B)Random Forest, (C) Logistic Regression and (D) XGBoost}
\label{fig:LIME_Collection_Outputs}
\end{figure*}

Running the code presents us with the LIME output, displayed in \Cref{fig:LIME}, consisting of three parts: the prediction probabilities on the left side, the feature probabilities in the center and the feature-value table on the right. The prediction probabilities graph shows the model's decision on that instance, meaning which outcome it predicts and the corresponding probability. In our example it displays the output of the logistic regression and predicts, that it is not going to rain with 83\% probability, represented by the blue bar with the number 0 and that it is going to rain with 17\%, represented by the orange bar with the number 1. The feature probabilities graph gives insight into how much a feature influences the given decision. For this observation the variable \textit{WindGustSpeed} is the most influential factor and supports the prediction, that it is not going to rain tomorrow. The second most important feature is \textit{Humidity3pm} which weights towards that it is going to rain tomorrow, represented by the number 1. In this case, we display the top five features in our output, but theoretically all the features could be listed that way, ordered by their importance. The last graph is the feature-value table, which also sorts the features by importance, but instead of showing their weight, is given the actual value that this feature possesses in this observation. For example, the fifth feature, \textit{MaxTemp}, shows 21.50 in this table, representing 21.5 degrees Celsius, the maximum temperature on the day of the observation. It is coloured blue, as it is influencing the model's decision towards no rain. As demonstrated in \Cref{fig:LIME_Collection_Outputs}, LIME does not differentiate between the machine learning model used but displays each of them the same way.

		\subsection{Evaluating the models on a global level }\label{sec:Quant} 
		
In an attempt to analyse the LIME output on a more global level, we apply the framework on fifty observations and aggregate the output in an excel file to compare the graphs with each other. As we analyse four models, we end up with 200 interpretations in total. 

LIME allows us to look at individual features in more detail and evaluate their influence, the occurrences of the three most relevant features are summarized in table \autoref{Tab:Quant_Summary}. In our analysis the framework displays the top five features per observation resulting in 200 total feature counts and 50 top positions per model. Out of this set, \textit{Humidity3pm} occurs most frequently, except for the XGBoost where it is ranked second after \textit{Pressure9am}. It appears 50 times in the analysis of the decision tree and logistic regression, 42 times at the random forest and 48 times at the XGBoost. Furthermore, \textit{Humidity3pm} is not only the most frequent, but is also considered the most important feature, as for the logistic regression it is the most influential feature, meaning it is ranked number one, in all 50 cases and for the decision tree in 42 cases. In case of the random forest, its prediction that it is not going to rain is heavily influenced by \texttt{Rainfall}, as whenever it did not rain, it is ranked in first or second position, which happens in 22 and 11 cases, respectively. Nevertheless, \textit{Humidity3pm} is also important for the random forest and occurs in 21 cases on the first rank. In the XGBoost classification \textit{Humidity3pm} is considered the most important feature 38 times. 
The least considered features are \textit{WindGustDir}, \textit{RainToday} and \textit{Temp9am}, with an occurrence of five, seven and eight times, respectively, none of which are ever ranked within the first or second position.
Considering this values, we now know that \textit{Humidity3pm} is highly predictive for our models, bringing us a step closer to developing a usable application.

\begin{table*}[t!]
\centering
\small
  \begin{tabular}{lcccccccc}
    \toprule
    \multirow{2}{*}{Feature} &
      \multicolumn{2}{c}{Decision Tree} &
      \multicolumn{2}{c}{Random Forest} &
      \multicolumn{2}{c}{Logistic Regression} &
      \multicolumn{2}{c}{XGBoost} \\
      & {O} & {TP} & {O} & {TP} & {O} & {TP} & {O} & {TP} \\
      \midrule
    Humidity3pm & 50 & 42 & 42 & 21 & 50 & 50 & 42 & 38 \\
    Pressure9am & 50 & 4 & 37 & 2 & 20 & 0 & 48 & 5 \\
    WindGustSpeed & 50 & 4 & 29 & 4 & 44 & 11 & 34 & 4 \\
    \bottomrule
  \end{tabular}
  \caption{Summary of the most occurring (O) and highest rated (TP) features}
  \label{Tab:Quant_Summary} 
\end{table*}

By displaying the intervals of its classification, LIME enables us to evaluate the accuracy of a single prediction. In terms of a false assessment we calculate the absolute difference between the probabilities assigned to the target variables, measured in percent. This tells us by how much the prediction is wrong and results in another indicator to assess the models. The false classifications are divided into two categories: a wrong prediction with less than 20 percent of absolute difference is called a close miss and a prediction with 20 percent or over more absolute difference is called a far miss. The results are displayed in \autoref{Tab:Misses}. The analysis of all observations results in the following: the decision tree classifies 12 out of 50 instances incorrectly, which are split evenly between close and far misses. The average absolute difference of all wrong classifications is 23 percent. In terms of the amount of incorrect classifications the logistic regression performs better than the decision tree, with eight wrong classifications, of which five are a close and three are a far miss. In absolute difference the logistic regression performs slightly worse, with around 26 percent. The random forest misclassifies nine times, of which six are close and three are far misses and gives us an average of 15 percent. Lastly, the XGBoost predicts incorrectly only four times, one time causing a close miss and three times a far miss, resulting in around 38 percent absolute difference, which is significantly lower in the times of incorrect classifications, but when it fails than by a lot more than other models. 

Considering the different evaluations we conducted, XGBoost is superior in the majority of cases. With the highest accuracy of 85\%, weighted classification report scores of 85\% \textit{precision}, 85\% \textit{recall},  84\% \textit{f1-score}, a \textit{ROC-test-baseline} of 88\% and the least amount of incorrect classifications, it delivers a better performance than the other models.

\begin{table*}[t!]
\centering
\small
  \begin{tabular}{lcccc}
    \toprule
    Type & DT & RF & LR & XGB \\ 
    \midrule
   Num. of close Misses (< 20\%) & 6 & 6 & 5 & 1 \\
    Num. of far Misses ($\geq$ 20\%) & 6 & 3 & 3 & 3 \\
    Average (in \%) & 23 & 15 & 26 & 38 \\
    \bottomrule
  \end{tabular}
  \caption{Summary of close and far misses}
  \label{Tab:Misses} 
\end{table*}


\section{Evaluating LIME from a usability perspective}\label{sec:Sec5} 


This usability assessment is split into two parts: firstly, we perform interviews to get an impression of how LIME is interpreted by people who are not familiar with the concept of explainable AI; secondly, we use a user experience evaluation framework in order to perform a self assessment of LIME's usability based on its criteria.

\subsection{The interviews} 
We interviewed six people, equally split between male and female, three with prior knowledge of machine learning and three with no prior knowledge. None of them were familiar with the concept of xAI before participating in the interview. The participants were either academics or in the process of pursuing a degree. In each interview we wanted to find out how interpretable the LIME output is for a person who never worked with xAI before. An overview of the interview results discussed herein is displayed in \Cref{Tab:Interview}. 

\begin{figure*}[t!]
\includegraphics[width=\textwidth]{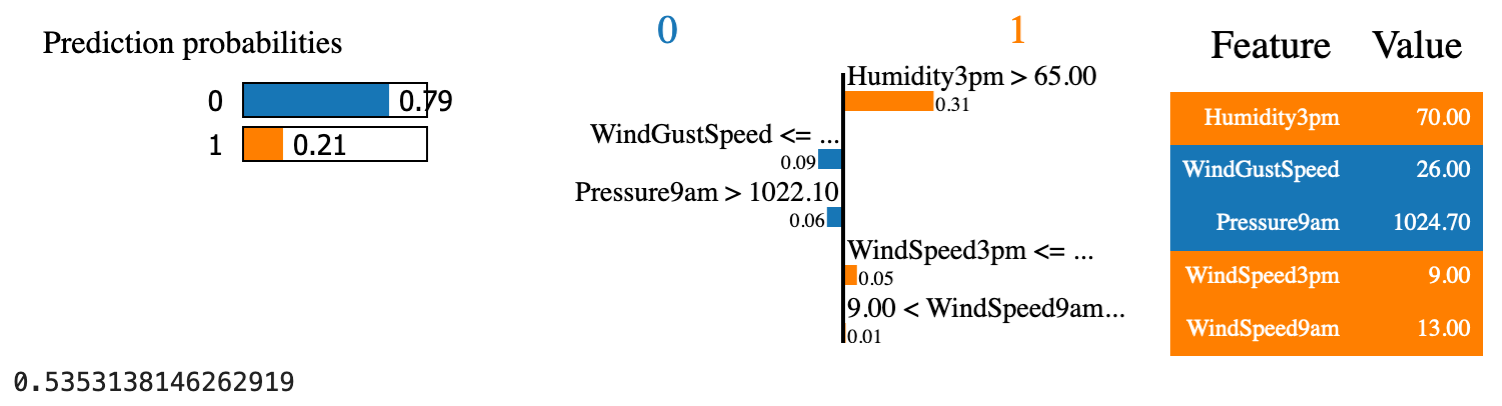}
\caption{Example of the interview LIME output}
\label{fig:LIMEOut}
\end{figure*}

The interview was split into two sections, both  of which started with an explanation from the interviewer. In the first part the interviewees were given a quick introduction into rain prediction, as well as a quick introduction into the applicable machine learning methods. They were subsequently shown the first \texttt{LIMETabularExplainer} output graph (cf., \Cref{fig:LIMEOut}) and were asked the following four questions.

\begin{description}[style=unboxed]

\item[\textit{What do you see in this graph?}]
All interviewees expressed uncertainty about what the illustrations show. All started with identifying the three graphs and tried to make sense of the different numbers. Although a few participants struggled with the prediction-probabilities and the feature-value graph, every participant had difficulties interpreting the feature probabilities as the numbers did not seem to add up and there was too much information given in a badly structured way.

\item[\textit{Which feature influences the prediction and how?}]
People without prior machine learning knowledge struggled to see the relation between the prediction probabilities and the classification, but those with prior knowledge in machine learning concluded, that there is a connection between the feature probabilities and the prediction probabilities graph. Two concluded correctly, that the second smaller numbers on the central graph are probabilities, as they are between 0 and 1 and influence the attribute of the predictability.

\item[\textit{Do you know why the model made this prediction?}]
Only one out of six answered correctly, that the classification is determined by the numbers of the feature probabilities graph.

\item[\textit{How well can you interpret the results of the prediction of the graph, on an increasing scale from 1-10?}]
The interpretability of the LIME output was rated with an average of 3.66. The rating between the subgroups differed only slightly, the participants without prior knowledge gave an average of 3.33 and the participants with prior knowledge 4.0, respectively.


\end{description}

\noindent The second section started with a short explanation of each graph of the LIME output as well as an explanation of the meaning of the r-squared value at the bottom of the output. The participants were subsequently shown another LIME output and were asked four more questions.

\begin{description}[style=unboxed]

\item[\textit{What do you see in the second graph?}]
After the participants were given the explanation for each graph the answers improved significantly. Four understood the graphs correctly, but were still uncertain where the probabilities of the prediction probabilities graph came from. Two of the participants with a machine learning background understood the framework after the explanation. Another four pointed out that the r-squared scores of both models were low, which resulted in concerns about the reliability of the prediction.

\item[\textit{How well can you interpret the results of the prediction, on an increasing scale from 1-10?}]
Even though several remarks were made in the previous question the interpretability of the graph after the explanation improved significantly, to an average of 7.66, while participants with prior machine learning knowledge again rated it slightly higher with an average of 8.16.

\item[\textit{What differences do you see between this one and the other graph?}]
All participants noted the different prediction probabilities. Some participants pointed out that there is a big difference on how the features in the different outputs were rated and that the numbers of the feature value graph had changed.

\item[\textit{Is there anything that stands out as strange or unusual?}]
Additionally, five out of six participants stated that the central graph was not very interpretable and three mentioned that they found the choice of colours disturbing. Furthermore, five interviewees suggested a legend, titles or a short explanation should be included in the output visualisation to improve its interpretability.

\end{description}

To sum up, the results produced by the framework are difficult to understand without documentation and/or explanation. Although the participants with a background in machine learning were more effective in terms of interpreting the explanation produced by LIME, usability assessments such as the one described in this paper could be used to significantly improve the user experience.

\begin{table*}[t!]
\centering
\small
\scalebox{1.0}{
\begin{tabular}{rccccrcrc}

\rot{Participant} &
\rot{ML knowledge}&
\rot{Gender}&
\rot{Illustration}&
\rot{Prediction}&
\rot{Rating part I}&
\rot{Understand part II }&
\rot{Rating part II}&
\rot{R2} \\ \toprule 
1      &    yes     &   m   &  yes  &   yes &   3   &  improved &  8    &     \\ \hline 
2      &    no      &   f   &  no   &   no  &   3   &  improved &  6.5  &  x    \\ \hline
3      &    no      &   m   &  no   &   no  &   4   &  improved &  7.5  &  x    \\ \hline
4      &    yes     &   m   &  yes  &   yes &   5   &  improved &  9.5  &     \\ \hline
5      &    no      &   f   &  no   &   no  &   4   &  improved &  7.5  &  x    \\ \hline
6      &    yes     &   f   &  no   &   no  &   3   &  improved &  7    &  x    \\ \bottomrule
\end{tabular}}
\caption{Participants' understanding of the LIME output}
\label{Tab:Interview} 
\end{table*}

    \subsection{Self assessment of the usability}

To assess LIME's user experience more broadly, we adopt the definition of usability proposed by the International Organisation for Standardisation (ISO)\footnote{https://www.iso.org/home.html} in their ISO 9241-11 1998 report \cite{ISO}. Therein, usability is defined as the \textit{"extent to which a system, product or service can be used by specified users to achieve specified goals with effectiveness, efficiency and satisfaction in a specified context of use"} \cite{ISO}. As this definition is too broad to be directly applied in our evaluation context, we improve its applicability by taking into consideration the \textit{"New ISO Standards for Usability, Usability Reports and Usability Measures"} produced by \citet{ISOextend} and  the \textit{"Usability Meanings and Interpretations in ISO Standards"} guidelines provided by Abran et al. \citet{Abran}.

\subsubsection{How effective is LIME in terms of achieving model interpretability?}
In terms of effectiveness, \citet{ISOextend} state that \textit{"effectiveness has been associated with completing a task completely and accurately, but it is also important to take account of the potential negative consequences if the task is not achieved correctly".} From this we extract three effectiveness factors: measure of completion; measure of accuracy; and negative consequences to rate effectiveness. \citet{Abran} take a more holistic perspective questioning \textit{"how well do users achieve their goal using the system?"}.
Thus, we use both the standard and the guidelines in order to develop four effectiveness questions tailored specifically to LIME, and subsequently use them to perform our assessment: 

\begin{description}[style=unboxed]

\item[\textit{(a) How complete is the explanation on a local level?}] 
LIME is a local explainability framework, therefore it calculates the influence of every feature and its importance on a local level (i.e., this is done for each prediction). Nevertheless, the connection between the prediction probabilities and the feature probability graph is incomplete as currently only the feature importance score is shown. Additionally, these scores do not add up to the prediction probabilities. As displayed in \Cref{fig:LIMEOut}, the feature \textit{Humidity3pm} with a feature probabilities score of 0.31 alone exceeds the total prediction probability of 0.21 that it is going to rain, while the overall classification was in favor of no rain. This can only be explained by assuming that the displayed prediction probabilities are not the sum of the feature probabilities, but the result of another calculation not obvious to a user.

\item[\textit{(b) How complete is the explanation on a global level?}]
While LIME is generally used for local interpretability, in this paper we also assess its performance on a global level. It is not surprising that the \texttt{LIMETabularExplainer} is less effective globally, as it does not include a function or interface to allow a global evaluation. Thus, it is necessary to extract several observation outputs manually and analyse them in an Excel file, as we did in the global analysis of \Cref{sec:Sec5}.
Considering the importance of global interpretability and the effectiveness of the simple proof of concept presented in this paper, it would be beneficial to: (i) implement performance indicators that allow for a global comparison with other models; and/or (ii) add a function to extract the local outputs of several random observations as a spreadsheet, so the user can calculate indicators necessary for a global comparison themselves.

\item[\textit{(c) Could accurate results be misinterpreted?}]
The interpretations of the local predictions appear to be accurate. But we see a risk of misinterpretation when it comes to the tabular explainer, as no comprehensive explanation of it has been published yet \cite{LIMEdocumentation}. Therefore, we have to rely on third party explanations like online articles\footnote{https://medium.com/analytics-vidhya/explain-your-model-with-lime-5a1a5867b423} \footnote{https://www.oreilly.com/content/introduction-to-local-interpretable-model-agnostic-explanations-lime/} or talks on YouTube\footnote{https://www.youtube.com/watch?v=CY3t11vuuOM}\footnote{https://www.youtube.com/watch?v=C80SQe16Rao}. Ideally such guidance should be incorporated into the LIME documentation.

\item[\textit{(d) What negative consequences arise from a misinterpretation?}]
In case of a misinterpretation of the LIME evaluation the severity of the negative consequences depends on the use-case. For example the implication of the predictions produced by our rain prediction model for Australia and an  automated defense system \cite{Gunning} differ greatly. In our case a mistake in the interpretation could lead to a faulty feature importance and therefore a wrong rain forecast. In the automated defense system case an incorrect classification could put lives at risk. As the severity of the consequences is not determined by the developers of LIME but rather lies in the hands of the users, reducing the risk that a misinterpretation occurs should be one of the key evaluation criteria when it comes to usability assessments.

\end{description}

\subsubsection{What resources are consumed in order to achieve interpretability?}
In order to evaluate resource efficiency Bevan et al. \citet{ISOextend} identify the following factors: task time, time efficiency, cost-effectiveness, productive time ratio, unnecessary actions and fatigue. We aggregate them to a list with mutually exclusive components and conclude with the question raised by Abran et al. \textit{"What resources are consumed in order to achieve the goal?"} \cite{Abran}. 

\begin{description}[style=unboxed]

\item[\textit{(a) How much time does it take to use LIME?}]
Both, the time to set up LIME as well as the time to analyse the output play a role in this context. The setup works well, however the official \texttt{LIMETabularExplainer} setup documentation relates to several old packages\footnote{https://lime-ml.readthedocs.io/en/latest/lime.html}. Therefore, the initial process of applying the original notebook and trying to find workarounds consumed a lot of time. %
Additionally, the analysis of the LIME output took a considerable amount of time, as the documentation of the graphs is non-transparent as stated in the effectiveness evaluation. On the up-side, the time it takes to compute and display an observation is minimal.

\item[\textit{(b) What other costs are involved?}]
As LIME is an open source tool, no licensing costs are involved and also the publications, documents and videos to understand the tool (where available)  are can be freely accessed.

\item[\textit{(c) Does this process cause fatigue?}]
Applying LIME to only a few observations can be performed quickly and therefore is not costly from a performance perspective. However, the global interpretation was a tedious process,  which entailed hours of repetitive manual work copying and pasting LIME output from the notebook into an Excel file. Also, given that there is no benchmark on the number of observations necessary to evaluate the models globally it is not clear how many outputs are necessary/sufficient.

\end{description}


\subsubsection{How satisfying is the application of LIME?}
Satisfaction is the least standardised of the three parameters as it is highly dependent on the user and use-case \cite{ISOextend}. Based on Bevan et al. satisfaction aims to take \textit{"positive attitudes, emotions and/or comfort resulting from use of a system, product or service"} \cite{ISOextend} into account. The question Abran et al. raise to assess satisfaction is \textit{"How well does the user feel about the use of the system?"} \cite{Abran}, which we include in our analysis. Combining both ideas we come up  with the following assessment questions:

\begin{description}[style=unboxed]

\item[\textit{(a) Do we have a positive or negative attitude towards the tool?}]
At the start of the implementation our attitude was very positive, as LIME's serves to help users to interpret and trust predictions performed by blackbox algorithms. During the setup our attitude deteriorated due to a lack of documentation and support, which posed an even bigger problem during the analysis. LIME gives insight into a model's processes, but here again it takes a lot of effort to get a clear understanding of the framework, which has a negative influence on our attitude. Naturally, once we learned how to apply and interpret LIME, the process was a much more pleasant one.

\item[\textit{(b) What emotions arise from using it?}]
The lack of clear and explicit guideline makes understanding LIME a frustrating process. However, reaching the point of a better overall understanding of our classification models raises positive feelings. Especially LIME's short processing time makes it easy to evaluate several instances in a row, which leads to a very pleasant user experience. 

\item[\textit{(c) How satisfying is the final result?}]
The output of the \texttt{LIMETabularExplainer} unquestionably helps to understand the model's classification process, as it offers insights conventional methods can not provide, which causes satisfaction. However, this satisfaction could be increased by eliminating doubt about the relationships between the local indicators and offering a global analysis.

\end{description}



\section{Conclusions}\label{Sec:Sec6} 

In this paper, we assessed the effectiveness of the Local Interpretable Model-Agnostic Explanations (LIME) xAI framework, with a focus on its performance in terms of making tabular models more interpretable. In particular, we examined the performance of four state of the art classification algorithms on a tabular dataset, used to predict rain. In order to get a better understanding of the perofrmance of our models we used the \texttt{LIMETabularExplainer} to analyse single observations on a local level, and merged several observations such that it was also possible to evaluate the models on a global level. In order to assess the interpretability of the output produced by LIME, we conducted interviews with individuals who have never worked with LIME. In addition, we developed a usability assessment framework, derived from the International Organisation for Standardisation 9241-11:1998 standard, in order to summarise our experience using LIME. Overall, we conclude that while LIME helps to increase model interpretability, usability studies are needed in order to improve the user experience, and tools and techniques are needed in order to facilitate global comparisons.

In terms of future work, besides proposing strategies for improving the interpretability of the output produced by LIME, and the usability of the framework from a global level perspective, we are interested in using our interpretability framework to benchmark alternative model-agnostic explanation frameworks.






\bibliographystyle{elsarticle-num-names}
\bibliography{manuscript.bib}







\end{document}